\documentclass[sigconf]{acmart}
\usepackage{multirow}
\usepackage{scalefnt}
\usepackage{smartdiagram}
\usepackage{tikz}
\usepackage{indentfirst}
\usetikzlibrary{shapes.geometric, arrows}
\makeatletter
\newif\if@restonecol
\makeatother

\usepackage[linesnumbered,ruled,vlined]{algorithm2e}%
\usepackage{algpseudocode}
\renewcommand\footnotetextcopyrightpermission[1]{} %
\AtBeginDocument{%
  \providecommand\BibTeX{{%
    \normalfont B\kern-0.5em{\scshape i\kern-0.25em b}\kern-0.8em\TeX}}}
\begin{document}
\title{Ymir: A Supervised Ensemble Framework for Multivariate Time Series Anomaly Detection}
\author{Zhanxiang Zhao}
\affiliation{%
  \institution{}
}
\email{}

\begin{abstract}
We proposed a multivariate time series anomaly detection framework Ymir, which leverages ensemble learning and supervised learning technology to efficiently learn and adapt to anomalies in real-world system applications. Ymir integrates several currently widely used unsupervised anomaly detection models through an ensemble learning method, and thus can provide robust frontal anomaly detection results in unsupervised scenarios. In a supervised setting, domain experts and system users discuss and provide labels (anomalous or not) for the training data, which reflects their anomaly detection criteria for the specific system. Ymir leverages the aforementioned unsupervised methods to extract rich and useful feature representations from the raw multivariate time series data, then combines the features and labels with a supervised classifier to do anomaly detection. We evaluated Ymir on internal multivariate time series datasets from large monitoring systems and achieved good anomaly detection performance.
\end{abstract}

\maketitle
\pagestyle{plain} %

\section{Introduction}

In recent years, anomaly detection of multivariate time series has been widely used in different domains, such as sensor devices, network services, and mickle complex system applications.

Although lots of unsupervised anomaly detection algorithms have been proposed for specific applications or detecting specific types of anomalies, none of them can achieve good anomaly detection performance across several industrial systems with different anomaly criteria. In real scenarios, we have to tackle massive time series, the abnormal patterns of which are quite complicated and different across systems. Therefore, it's hard for a single unsupervised method to meet all the underlying assumptions about anomalies in different systems, which leads to poor anomaly detection performance.

\begin{figure}
    \centering
    \includegraphics[width=\linewidth]{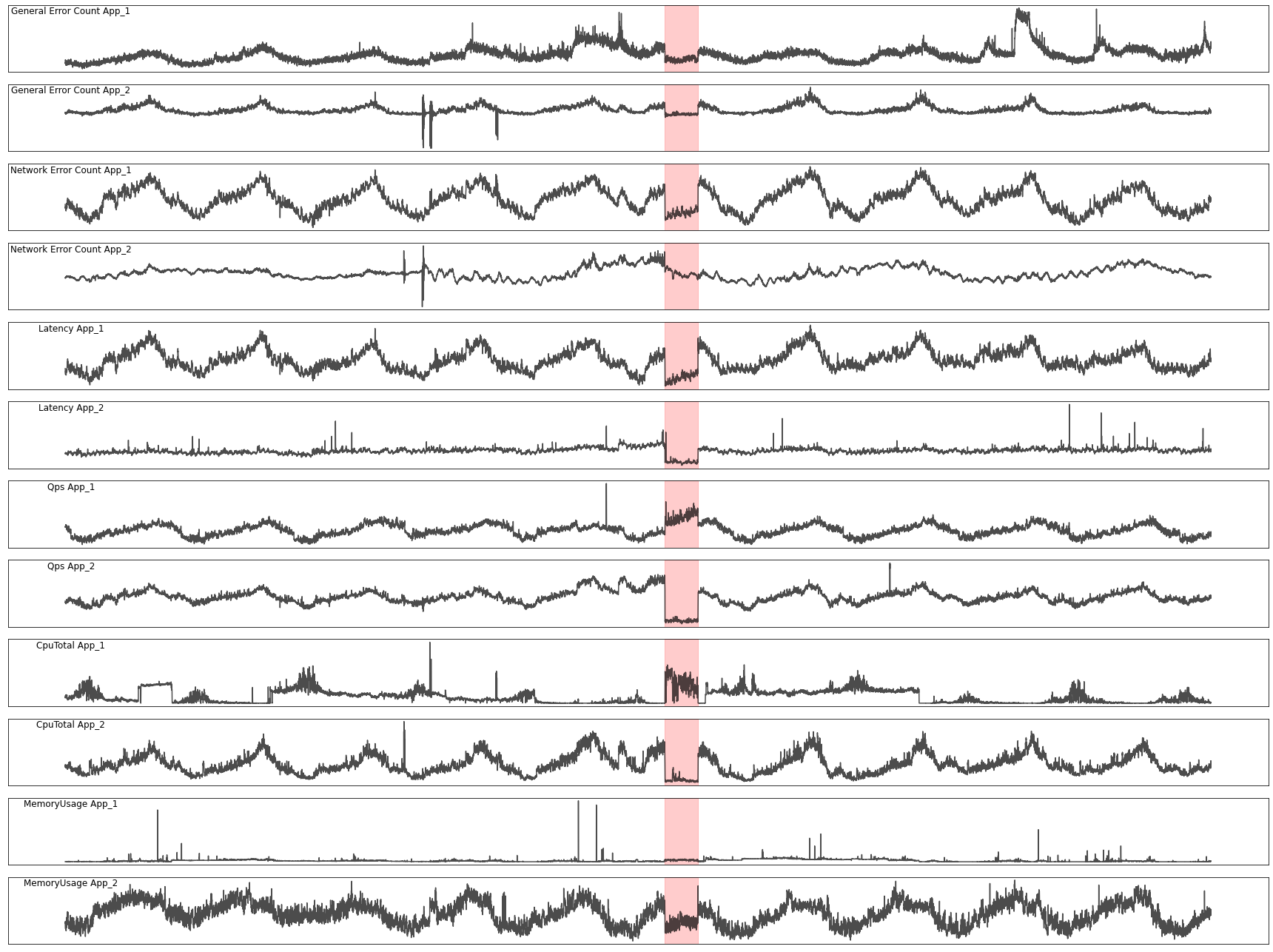}
    \caption{False Anomaly: Restart Server}
    \label{fig:exp}
\end{figure}

In addition, we found that the detection results obtained by the unsupervised anomaly detection models often do not meet the system users' interests about anomalies.
For example, as shown in Figure \ref{fig:exp}, in a multivariate time series collected from a system application, the simultaneous drop in indicators such as CPU, Memory, QPS may only be due to a result of the restart of the server. Such patterns are often judged as anomalies by unsupervised anomaly detection models, but system users will only ignore it. Therefore, it's important to combine the feedback provided by users into the anomaly detection model, which can help the algorithm automatically learn and adapt to users' interests about anomalies and improve anomaly detection performance.

In this paper, we propose an ensemble learning framework that integrates a variety of unsupervised anomaly detection models with different characteristics. We then combine the ensemble features with user feedback using a supervised classifier model to achieve high anomaly detection performance in real industrial systems.

\section{Related work}\label{related_work}
Based on the advantages of ensemble learning, XGBOD ~\cite{zhao2018xgbod} uses a variety of unsupervised anomaly detection algorithms to extract rich features from the original data, and then leverages XGBoost to learn the extracted features in a supervised setting. Ymir's framework is similar to XGBOD, and both use unsupervised models to extract decision boundaries and apply them to supervised models.
\section{Methodology}\label{methodology}
In this section, we describe each component of Ymir in detail. We first introduce the framework of Ymir, followed by the structure of unsupervised representation models, the unsupervised detection model, and the supervised detection model.

\begin{figure}
    \centering
    \includegraphics[width=\linewidth]{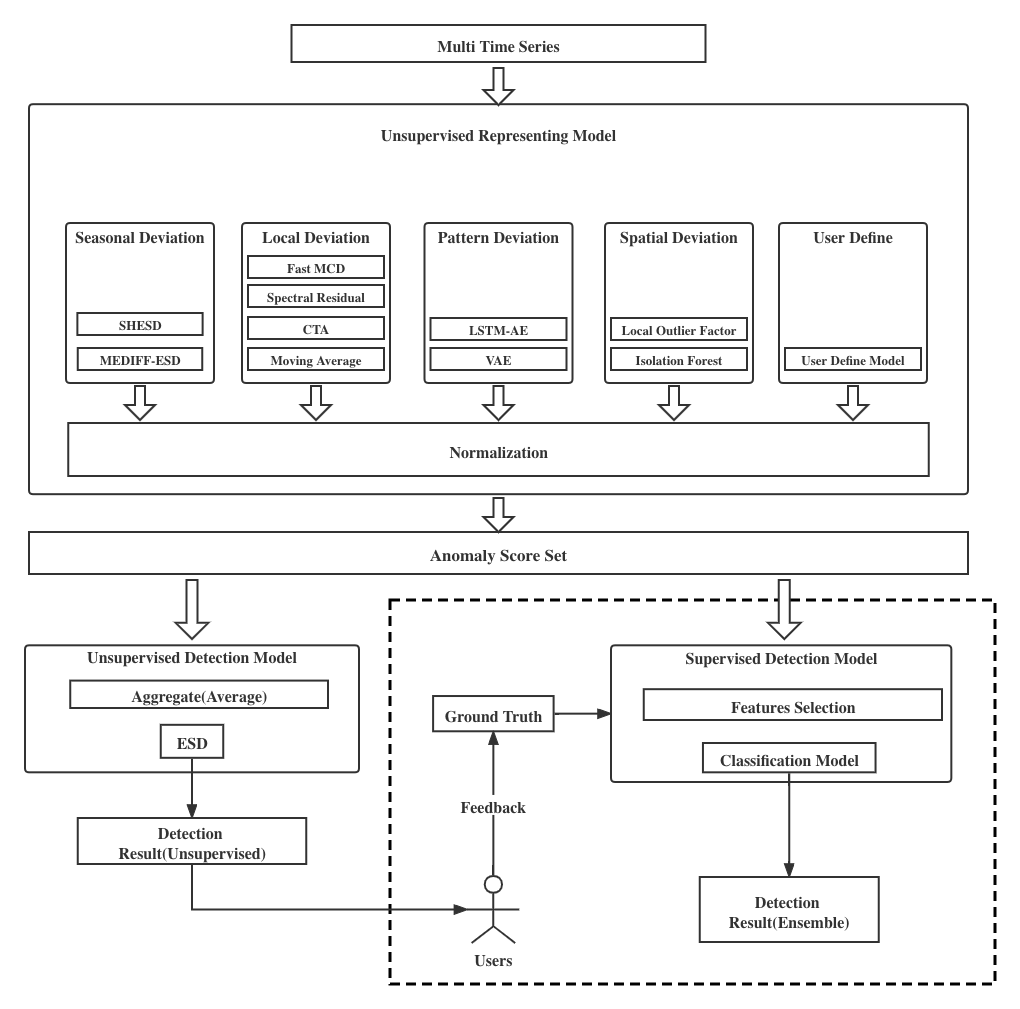}
    \caption{Ymir Framework}
    \label{fig:arch}
\end{figure}
The framework of Ymir is shown in Figure \ref{fig:arch}. The framework can be divided into two parts. The first part is an unsupervised representation module, which is used to extract and normalize the decision boundary of each model. The second part is the anomaly detection module. In unsupervised and supervised scenarios, anomaly detection is performed on the normalized anomaly scores based on the ESD ~\cite{ESD} method and supervised classifier, respectively.

\subsection{Unsupervised Representation Module}
\label{feature-extraction}
There are various abnormal patterns in multivariate time series, and thus we need several anomaly detection models with different properties to characterize the anomalies in the original data. We divide the anomalies in our system into 5 categories based on domain knowledge, including periodic anomalies, local anomalies, change points, spatial anomalies, and user-defined anomalies. We leverage statistical methods such as MEDIFF ~\cite{mediff} and SHESD~\cite{shesd} to capture periodic anomalies.
We employ moving average, Chebyshev theory ~\cite{cta}, spectral residuals, etc, to capture local anomalies. For change points anomaly, we apply deep models like VAE~\cite{vae} and LSTM-AE~\cite{lstmae}. We also use Isolation Forest~\cite{isof} and LOF\cite{lof} to capture spatial anomalies. Moreover, Ymir allows users to customize anomaly detection functions for user-defined anomalies.

Given an input $TS$ consists of $n$ time series $TS = [ts_1, ts_2, \dots, ts_n]$, the length of each time series are the same. Then we select $k$ anomaly detection models, denoted as $M = [m_1, m_2, \dots, m_k]$. The $k$ anomaly detection models respectively detect the original $n$ time series and output the decision boundaries, denoted as $B = [b_1, b_2, \dots, b_k]$. Then we normalize $B$ into [0, 1] and use it as feature scores $AS = [as_1, as_2, \dots, as_k]$. 
Finally, the feature scores $AS$ can be directly used to calculate the anomaly scores in unsupervised anomaly detection, or used as features to train the classifier in the supervised setting.

\subsection{Anomaly Detection Module}
In unsupervised setting, we directly perform a weighted average of $AS$ mentioned in section \ref{feature-extraction}. $W = [w_1, w_2, \dots, w_k]$ are the weights for each anomaly score in $AS$, which represent the users' preference on different types of anomalies. The default weight is 1 for each feature, and can be customized by system users. Then We obtain the weighted anomaly score $AS_{weighted}$, denoted as $AS_{weighted} = [w_1*as_1, w_2*as_2, \dots, w_k*as_k]$, and we directly perform the ESD on $AS_{weighted}$ to get the unsupervised anomaly detection results, denoted as $Result_{base}$.

\begin{figure}
    \centering
    \includegraphics[width=\linewidth]{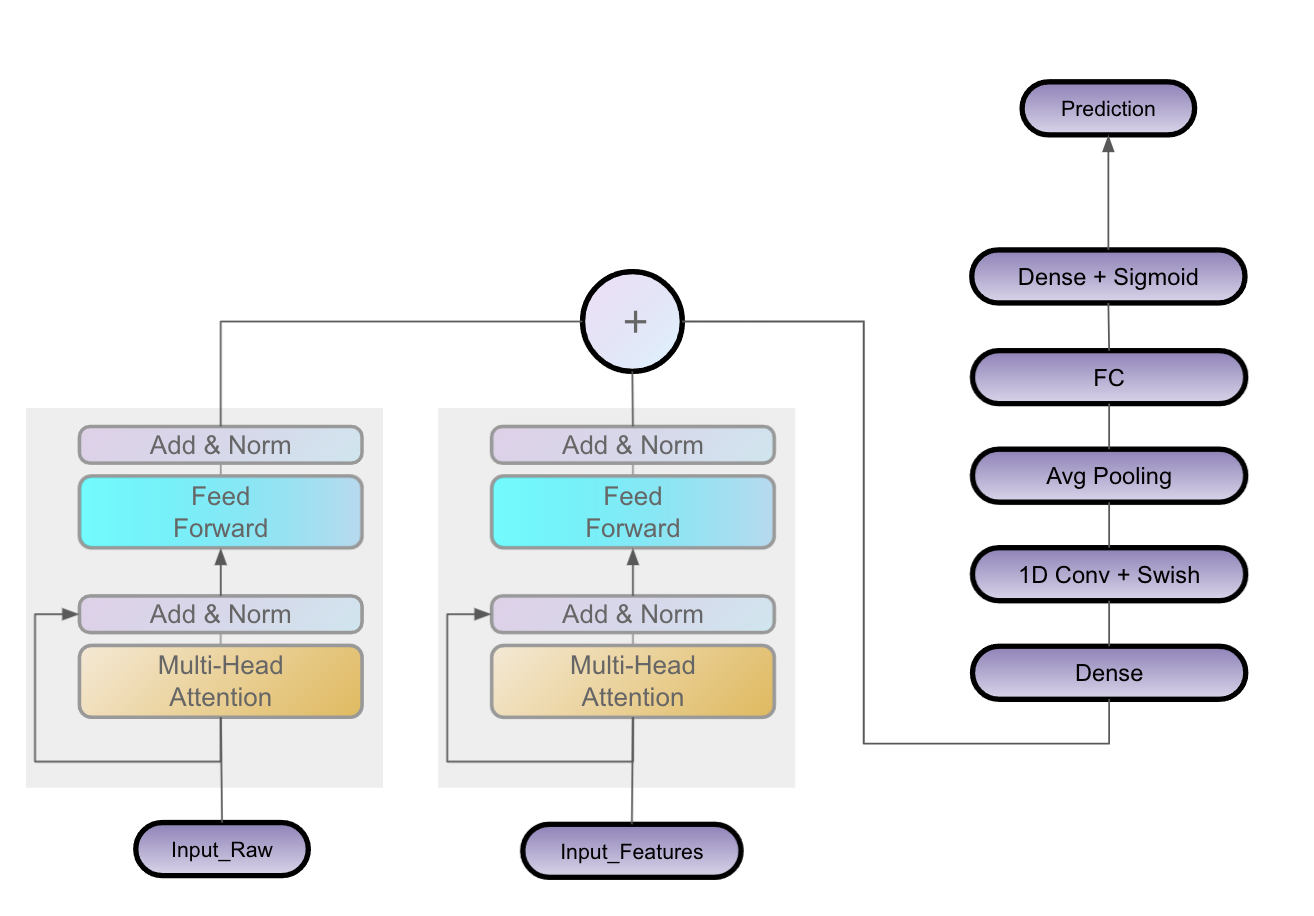}
    \caption{Supervised Classification Model}
    \label{fig:model}
\end{figure}

In case partial or full feedback labels are provided by domain experts and system users, we replace the ESD with a classification model for anomaly detection. The input of the classification model consists of two parts, the original time series $TS$ and the aforementioned feature scores $AS$. The structure of the classifier is shown in Figure \ref{fig:model}. The classifier mainly contains two Transformer~\cite{transformer} modules and a 1DCNN module. We first employ the two Transformer modules to learn the main components from original data and abnormal decision boundaries, then concatenate the embedding outputs of the two Transformer modules. Finally, the 1DCNN compresses the embedding and output prediction results.

Specifically, for a fully-supervised setting, we can directly use the full labels for training the classifier. 
However, when only a few labels are provided, the classifier cannot be well-trained with limited data and labels, which leads to poor anomaly detection performance. Therefore,
we train a semi-supervised Ymir as follows: We first set a confidence threshold $th$, and use the aforementioned $Result_{base}$ with a confidence higher than $th$ as pseudo abnormal labels, while the rest are pseudo normal labels. The pseudo label is denoted as $Label_{base}$. Then we use the few user feedback labels to replace the pseudo labels of corresponding points in $Label_{base}$ and obtain the final labels used for training the classifier. Specifically, in this scenario, we will have to adjust label-smoothing~\cite{labelsmoothing} based on the proportion of user labels to obtain more stable results.

\subsection{Online Detection and Experimental Results}

For online detection, Ymir first extracts features from the raw multivariate time series data using the trained unsupervised feature representation models. Then the raw multivariate time series and extracted feature scores are fed into the classifier model to get the detection results. With the help of 1.5 to 3 months of fully-labeled training data, the supervised ensemble Ymir is able to achieve 0.65 to 0.97 best range F1 score~\cite{tatbul2018precision} on the internal datasets extracted from different systems in our company, which significantly outperforms the state-of-the-art unsupervised multivariate time series anomaly detection methods.

\section{Conclusion}\label{conclusion}
In this paper, we proposed Ymir, a supervised ensemble framework for multivariate time series anomaly detection in monitoring systems. 
According to domain knowledge about the anomalies in industrial systems, Ymir integrates a variety of unsupervised anomaly detection models, which can extract features about different types of anomalies and obtain basic anomaly detection results in unsupervised scenarios. With the help of 1.5 to 3 months of user feedback labels, the supervised Ymir achieves a good anomaly detection performance on the multivariate time series data from different industrial systems. However, it is often hard to continuously obtain the full labels in real-world scenarios, and the performance of semi-supervised Ymir will drop as the feedback labels become fewer. Therefore, our future work is to improve Ymir's anomaly detection performance with few labels and other useful information.

\bibliographystyle{ACM-Reference-Format}
\bibliography{reference}

\end{document}